# Data-driven models and computational tools for neurolinguistics: a language technology perspective

Ekaterina Artemova[1], Amir Bakarov[1], Aleksey Artemov[2], Evgeny Burnaev[2], Maxim Sharaev[2]

*[1]National Research University Higher School of Economics, Moscow, Russia*
*[2]Skolkovo Institute of Science and Technology, Moscow, Russia*
*{echernyak, abakarov}@hse.ru, {a.artemov, e.burnaev, m.sharaev}@skoltech.ru*

## Abstract

In this paper, our focus is the connection and influence of language technologies on the research in neurolinguistics. We present a review of brain imaging-based neurolinguistic studies with a focus on the natural language representations, such as word embeddings and pre-trained language models. Mutual enrichment of neurolinguistics and language technologies leads to development of brain-aware natural language representations. The importance of this research area is emphasized by medical applications.

## 1. Introduction

Contemporary scientific research is unimaginable without powerful data processing tools offered by computational sciences, in particular, by data science and machine learning. Data-driven approaches, aiming to automatically establish dependencies among measurable quantities using algorithmic and optimization techniques, substantially extend the potential scope of modelled variables (ultimately targeting the explanation of all experimental observables). Modern neuroscience, in particular, neuroimaging studies, has greatly benefitted from incorporating data-driven approaches (Sejnowski et al., 2014; Vu et al., 2018). Neurolinguistics is the focus for this survey: the major goals of this paper are to analyse research projects that address such topics as brain mapping, neural decoding and characterisation of neural processes as well as their achievements and connections to natural language processing.

   Decades ago, it was thought that a single specific brain region is responsible for performing particular linguistic functions. Now the common idea is that brain connectivity is more important than activations in separate brain areas. Collecting neuroimaging data (e.g. EEG, fMRI, fNIRS, eye-tracking) during medical neurolinguistics studies helps to shed light on the causes of impairments suffered by language-related brain structures. Modern findings from both brain mapping and linguistics begin to make predictions about the breakdown in the brain function and its spatial location which is responsible for these impairments (Oota et al., 2018). The range of neurological and psychiatric disorders studied with both brain mapping and linguistic approaches is huge: it includes, among others, Mild Cognitive Impairment (MCI), aphasia (McKinney-Bock and Bedrick, 2019; Wise et al., 1991), Alzheimer′s disease (Pakhomov et al., 2016), focal brain lesions (Troyer et al., 1998), and schizophrenia (Robert et al., 1998). In less than the two decades of research with the imaging and data-driven methods, neurolinguistics has made leaps from the early attempts on relating separate words with the imaging data (Friederici et al., 2000; Fiebach & Friederici, 2004), to the whole-brain mapping from the natural stimuli (Huth et al., 2016), to decoding of the linguistic meaning from the brain activation data (Pereira et al., 2018).

   A central topic in neurolinguistics is the study of the relations between features of the presented stimuli and the corresponding brain responses. One commonly aims at capturing such dependencies for the broadest possible set of stimuli, ultimately arriving at a universal structural and functional description of the brain's capabilities related to language processing and representing the lexical information. As such, a core problem in neurolinguistics is the construction of an efficient semantic space that could represent most stimuli; to this end, particularly attractive tools for representing both spoken and written language are offered by the recent breakthroughs in Natural Language Processing (NLP). NLP uses large-scale computational models such as Distributional Semantics Models (DSMs) and computes real-valued vectors (commonly referred to as *word embeddings*) to represent the meaning of words; these representations have been convincingly demonstrated to perform exceptionally well on a range of real-world applications. However, DSMs show success not only in applied NLP tasks but also in experimental psychological studies (Landauer & Dumais, 1997), as word embeddings are assumed to represent a trustworthy model of semantics. The hypothesis that the word embeddings reflect semantics as it is proposed in the human cognition inspired the idea of utilising DSMs for neuroscience studies, such as the brain mapping: word embeddings act as a linguistically grounded word representation which could be leveraged to explore the related patterns in the brain.

Despite NLP-related methods being increasingly used in neuroimaging studies, to date, no survey exists that would cover the use of language technologies in the context of neurolinguistics. In the present work, we are endeavoring to shorten this gap by investigating the involvement of NLP-based methods for stimuli representation in a range of tasks in neurolinguistics studies. In particular, we consider brain mapping, brain decoding, and related tasks, and conduct a literature survey. We note, however, that we focus purposefully on the data-driven models, methods, and computational tools used in the context of neuro-linguistic tasks, as opposed to studying the actual goals of neurolinguistics itself. As such, our literature survey is centered around the state-of-the-art research thematically relevant to data science, machine learning, and language technologies, rather than neurolinguistics. Specifically, we review publications appearing recently in a number of journals and proceedings on NLP and machine learning, attempting to reveal the impact of data science and NLP technologies on the effectiveness of solving common neurolinguistics tasks. Where possible, we also attempt to correlate the motivation for the development of data-driven methods with the needs of applications.

This paper is structured as follows. Section 2 reviews the common experimental setups for the brain imaging-based neurolinguistics studies and includes discussion of the data acquisition issues as well as the addressed tasks, their respective challenges, and pre-built research environments. Section 3 introduces the major concepts of distributional semantics as well as an overview of the three generations of DSMs, while Section 4 presents brain mapping methods relying on the distributional semantics-based representations from their different generations. Section 5 introduces a recent trend of the brain-aware DSMs. Section 6 lists a few medical applications of the brain mapping and neurolinguistics. Section 7 concludes the article with a brief future outlook.

## 2. Experimental setup of neurolinguistics experiments

A common neuroimaging modality, utilised with the semantic mapping studies is the functional magnetic resonance imaging (functional MRI, fMRI), which aims to record the whole-brain, blood-oxygen-level-dependent (BOLD) responses, in human subjects, while they are exposed to the various types of stimuli. BOLD responses are commonly represented as dense 4-D arrays of the measured data, where time series of the blood flow-related activity measured in tens of thousands of *voxels* (which are small areas of size equal to approximately 2x2x2 mm3) are measured across the brain. These excitations are hypothesized to be elicited primarily by the presented stimulus (with a minor background contamination due to respiration, heartbeat, or movement).

Brain mapping aims to build interpretable models for the recorded BOLD data using tools common for statistics, machine learning, and other data science areas. When performing such modeling, commonly adopted premises are the following:

1)    Particular areas selectively respond to the particular content encountered in the stimulus. The stronger the influence, the higher the magnitude of the response. This suggests that we use simple linear models (for instance, the generalised and regularised linear regression) to determine which features of the stimulus lead to a high magnitude signal in each area.

2) Each area, represented by a voxel, responds largely independently of the other areas, thus a separate model is needed to fit responses in each cortical voxel.

Thus, one common path would be the voxel-wise modelling with linear models, which aims to estimate the influence of the content-related features found in stimuli on the fMRI measurements captured by the scanner. Studies aiming to build the brain mapping models differ by (1) the type of stimulus involved, (2) the feature representation chosen to describe the stimulus.

Interestingly, studies involving modern mathematical and computational tools have been recently involving a feedback loop, i.e. where an improvement to the computational models employed is sought. Such models must be predictive, i.e. one should be able to approximate the unseen brain activity from the newly presented linguistic stimuli. Another feature is that such models are hypothesised to be the good decoders of a linguistic meaning from the brain data, i.e. one may use them to extract a representation of meaning (and possibly decode it into text), based on the current brain signal.

Performing linguistic experiments with brain imaging is tricky due to many physiological processes, which take place in the brain during speech generation and comprehension. These processes highly depend on the experimental settings, so the majority of studies try to concentrate on a particular effect with strong control for others (Wehbe et al., 2014a). Brain activity is monitored during many kinds of audio and visual stimuli, ranging from more simple visual stimuli such as written words, in contrast to a non-word letter string (Salmelin, 2007), or a sentence with expected versus unexpected meaning (Kuperberg, 2007), to more complex audio and audio-visual stimuli like natural speech comprehension and movie watching (Hanke et al., 2014; Huth et al., 2016). One of the main research questions is the coordinated work, or integration of the so-called brain language network, including Broca′s and Wernicke′s brain, as well as the complicated information flow between them (Wehbe et al., 2014a). Another complication is the high dimensionality of the measured brain activity, the correlational nature of neuroimaging techniques, as well as the coarse measurements which the neuroimaging methods could provide (Abnar et al., 2019). For example, the dataset by (Wehbe et al., 2014a) consists of the fMRI scans of participants reading a particular text. The story was presented to the participants word by word on a screen, each word was displayed for 0.5 seconds and the fMRI scan was taken every 2 seconds, which is the usual temporal resolution for the fMRI technique (Abnar et al., 2019). It is also important to consider the physiological delay between the stimulus presentation and the brain haemodynamic response (Buckner, 1998) which could be around 4 to 6 seconds.

Moreover, as the raw fMRI signal is very noisy (Murphy et al., 2013; Sharaev et al., 2018a), and the noise could be highly correlated with the stimuli presentation, the former should be removed before the analysis (Sharaev et al., 2018b). In a well-known, 7T fMRI dataset, see (Hanke et al., 2014) participants listened to a German audio-description of the movie. During the fMRI scanning, the MR machine and the radiofrequency coil emit high and low frequency noise, which could be harmful for the speech comprehension. To avoid this effect, the audio signal was processed by a series of filters to remove the frequencies which would have caused acoustic distortions in the headphones (Hanke et al., 2014).

While fMRI has a good spatial (~1-3mm) resolution, its main disadvantage is a low temporal resolution (around 1-2 seconds), which prevents the registration of the rapid neuronal activity during speech comprehension and generation. When the stimulus is read, it usually takes around 100ms for the visual input to reach the visual cortex, 50ms to be processed as letter strings in a specialised brain region (Salmelin, 2007) and between 200-500ms, to have its semantic properties processed (Wehbe et al., 2014b). Magnetoencephalography (MEG) measures the small magnetic fields outside the skull, evoked by neuronal electrical activity, which is similar to electroencephalography (EEG). Thus, MEG recordings, like EEG, are directly related to neural activity and have no physiological delay. MEG data has a good temporal resolution (sampling frequency around 1kHz), which is perfect for tracking the fast dynamics of language processing. That's why this technique was used in some neurolinguistics studies (Wehbe et al., 2014b). The drawback of MEG is a poor spatial resolution, which prevents the precise activation localisation in the brain (Huster et al., 2012).

## 3. Three generations of distributional semantics

Distributional semantics models (DSM) stand for a group of methods that are used to map words from a large vocabulary to vectors. These vectors should consist of real numbers, have few zeros, and have relatively small dimensionality (in particular, it is common to construct vectors of dimensionality 300). Such word vectors are commonly referred to as word embeddings. These vectors are treated as mathematical (algebraic) objects: not only the similarity (or distance), between them, can be computed, but they can also be added or subtracted. At the core of numerous methods for computing word embedding is the distributional hypothesis: words that occur in the same contexts tend to have similar meanings (Harris, 1954).

Word embedding models are trained on large text corpora. They aim to find words that share contexts and represent them with such vectors that would be close, according to a mathematical similarity measure. For example[1], the embeddings of such words as *coffee* and *tea* should have a high similarity degree, since they are used in a similar way, along with the words *to drink, cup, to pour*, etc; a pair of words that occur together are called co-occurrence. What is more, advanced word embedding models allow us to conduct arithmetic operations: *coffee* to *morning* = *tea* to *evening*; *Madrid* to *Spain* = *Berlin* to *Germany*. Of course, these associations are corpus-specific and may not be present in other models.

From a historical point of view, there are three generations of DSMs: count-based DSMs, distributed or prediction-based DSMs, and deep contextualized DSMs (often named universal language models). Count-based DSMs arose in the early '90s, followed in the 2010s by prediction-based models. The recent breakthrough of deep contextualized DSMs leverages transfer learning techniques for downstream tasks.

---

[1] The examples are provided by RusVectores (https://rusvectores.org).

## 3.1 Count-based DSMs

Count-based DSMs build word representations by counting the usage of words in different contexts. The basic idea of such algorithms is that, given a sufficiently large number of texts, one could create a vocabulary of unique words and count the contexts of each word. By counting these contexts, one can represent each word through a real-valued vector.

To capture this co-occurrence, the so-called word-matrix is introduced. In this matrix, each unique word in a corpus is associated with an array of values. Each cell in this array is associated with another word that could be encountered in the context of this word, so basically the number of values is equal to the number of unique words in the corpus. Then, the value in each cell reports the degree of closeness of word $a$ to word $b$ from the context perspective. In the simplest case, it could be a binary value that says whether this word was encountered in the context of the considered word, or not. It could also be the number of encounters of this word, or more complicated measures which should, more precisely, report the "co-occurrence score", for example, Positive Pointwise Mutual Information. As a result, these actually count the word contexts, so they are referred to as count-based DSMs.

One of the most popular count-based models is Global Vectors (GloVe), presented by (Pennington et al. 2014). GloVe is a log-bilinear regression model, trained on aggregated global word-word co-occurrence statistics from a corpus, which factorizes the logarithm of the co-occurrence matrix.

## 3.2 Prediction-based DSMs

The second generation of DSMs are prediction-based models. These models leverage a language-modelling approach, by either predicting the current word from a context, or by predicting the context of the given word. Word2vec (word-to-vector) (Mikolov et. al, 2013), is a family of neural network architectures, namely Continuous Bag of Words (CBOW) and Skip-gram. See Fig. 1 for their illustration.

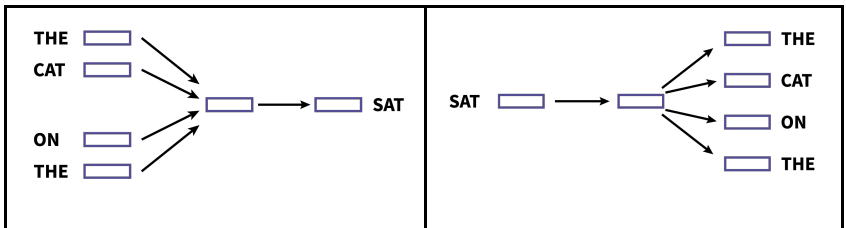

Fig 1. Word2vec configurations. (On the left): CBOW predicts a word based on its context, (on the right): skip-gram predicts the context of the given words.

A Word2vec model should be trained on a large text corpus. First, this corpus is preprocessed, so that each text is split in overlapping n-grams by a sliding window. Next, a neural network is trained using one of the Word2vec objectives, CBOW or skip-gram. The weights of the hidden layer are the actual learned word embeddings. The SGNS architecture is widely reused in other works. Fasttext (Bojanowski et al.) extends word2vec by producing word embeddings not for single tokens, but rather for character n-grams. Theoretically, this enables inference of an embedding for any unknown (out-of-vocabulary, OOV) word.

Lexical Vectors, LexVec, (Salle et al., 2016) combines the best of both worlds, namely, count-based and prediction-based models. This model inherits SGNS and negative sampling from word2vec, but unlike word2vec, which factorizes PMI matrix implicitly, it factorizes PPMI matrix explicitly.

Word embeddings can be manipulated as any mathematical object: not only is it possible to calculate a similarity between them, but also to sum them up or to subtract them. Word embeddings can be seen as a shallow representation of the grammar and semantics. When trained on a large general corpus, such as Wikipedia, word embeddings capture lexical semantics through context similarities – words that have similar meanings share similar contexts – so they can be revealed when measuring word similarity: for instance, vectors, corresponding to different forms of a single word, would be very similar, like "cat" and "cats", or vectors corresponding to synonyms, like "mug" and "cup", -- so the closer the words are, by lexical meaning, the closer the corresponding word embeddings should be. A similarity function of choice is usually a cosine function.

There is no straight linguistic interpretation of this similarity measure because it reflects different types of relations in different cases and is always affected by the parameters of the model. One of the most significant parameters is the training corpus, and the impact of the training corpus on the interpretation of vector similarity is still an open question. In most cases, this similarity reports the association (or co-hyponymy[2] of two words), in some cases, it reports relatedness[3]. The embedding model can be fine-tuned to express a certain type of similarity, such as meronymy[4] or hyponymy[5]. This approach to fine-tuning is referred to as retrofitting (Faruqui et al., 2015). The retrofitting approach has also been applied to construct not only more predictive but also more interpretable embeddings (Jauhar et al., 2015).

Word embedding models often fail when faced with such complex language phenomena as opposition, polysemy or hyponymy. Although word embeddings are exceptionally powerful for finding words that share a similar meaning, they are often mistaken for words that have opposite meanings[6], as they occur in similar contexts (Deriu, 2017). Word embedding models suffer from polysemy and homonymy. Such words, like "bank" or "apple", get a single vector, despite having multiple senses. A few models, such as AdaGram (Adaptive Skip-gram) (Bartunov et al. 2016) or SenseGram (Sense Skip-gram) (Pelevina et al. 2016), try to overcome this issue by simultaneous word sense disambiguation, and word embedding training.

---

[2] Such word pairs as "blue" and "red" or "tea" and "coffee" are co-hyponyms.

[3] Such word pairs as "car" and "gasoline" or "cat" and "striped" are related.

[4] "Tire" is a meronym of "car". Meronymy stands for "part of" relation.

[5] "Grey" is a hyponym of "color". Hymonymy stands for "is-a" relation. The inverse relation is hypernymy.

[6] Such words as "win" and "loose" have opposite meanings.

## 3.3 Deep contextualized DSMs

A natural extension of the concept of word embeddings is sentence embeddings. Sentence embeddings are used frequently in those applications, which require modeling of the sentence similarity. This could be utilised in such applications as paraphrase identification (whether one needs to identify degree of similarity of two sentences) or information retrieval (to find in a set of sentences those ones who are most similar to the queried sentence, and to rank them depending on their similarity). Paragraph2vec (frequently addressed as doc2vec, document-to-vector, (Le and Mikolov, 2014)) follows the architecture of word2vec to build a sentence embedding. The Skip-thoughts (Kiros et al., 2015) model is similar to skip-gram, too. It consists of one encoder, which encodes a sentence, and two decoders, such that one is used to generate the previous sentence, and the other is used to generate the next sentence. A skip-thought like task, combined with a conversational input-response and multiple classification tasks, are used to train the Universal Sentence Encoder, USE (Cer et al., 2018), by Google. Unlike Skip-thoughts, which is built from Long Short-Term Memory networks (LSTMs), USE exploits deep averaging networks and transformer-derived encoders. The majority of the aforementioned sentence embedding models suffer from several disadvantages: they are slow to train, give poor results or are difficult to use in real-life settings (Eger et al., 2019; Goldberg, 2019; Liu et al. 2019).

However, recent universal language models[7] (LMs) are a much more efficient solution to all these issues, as they search for context-dependent word embeddings, which are core to the transfer learning paradigm. The transfer learning paradigm is based on the idea of the pre-training of the large language model, on large corpora, and further fine-tuning of these models for downstream tasks. In NLP, one can think that, during the pre-training stage, the model learns the language. During the fine-tuning stage, the model learns to conduct specific tasks.   A new generation of DSMs comes with large, universal LMs, and they start to make a difference in every natural language processing task. Instead of the normal use of pre-trained word embeddings and training the whole model from scratch, now, a pre-trained language model is fine-tuned for downstream tasks. Word embeddings are an imperfect way to store language representation, which suffers from language ambiguity. Universal language models are less prone to polysemy and antonymy and are able to handle multi-linguality at the same time (Devlen et al., 2018, Pires et al., 2019).

---

[7] Such models are referred to as deep contextualized DSMs, pre-trained universal language models, and transformer-based language models interchangeably.

Inside the vast majority of transfer learning models are transformer layers, which are more advanced from a technical point of view, when compared to other layers. The transformer layer is essentially a feedforward network with an addition of a self-attention mechanism, which models the interactions of inputs. The self-attention mechanism shows how important a context of one word is to another word. In general, the architecture of transfer learning models is sophisticated, enumerates millions of parameters and takes weeks to be pre-trained. Even though pre-training of a large model is expensive and time-consuming, new models appear almost every month, as of late 2019. Among others, ELMo (Peters et al., 2018) and BERT by Google (Devlin et al. 2019) are the most popular models. In contrast to other models, ELMo is based on recurrent neural networks, while BERT and other models are transformer-derived. BERT's successors, ALBERT (Lan et al., 2019), RoBERTa (Liu et al., 2019), and other transformer-based models, Transformer-XL (Dai et al., 2019), XLNet (Yang et al., 2019), T5 (Raffel et al., 2019), released by Facebook, Microsoft and other technological companies, are larger and outperform BERT by a significant margin. ELMo, BERT and other models are trained in an unsupervised way by using language modelling objective or similar objectives.

ELMo exploits language modelling straightforwardly, while BERT's objective is combined from masked language modelling and next sentence prediction. Recent works have questioned the necessity of the next sentence prediction loss. RoBERTa, trained without the NSP loss matches or slightly improves downstream task performance of BERT. Transformer-XL is specifically designed to process long sequences by exploiting a segment-level recurrence mechanism. In contrast to BERT, which can process only a few sentences at a time, Transformer-XL is capable of modelling paragraphs. Universal language models can be used either as sentence embeddings by producing vector representation for an input text or a sequence of vector representations for the text tokens. In the case of the latter it is common to denote word vectors as context-aware word embeddings, as each word embedding attends the context of the word.

The General Language Understanding Evaluation (GLUE) benchmark, designed by (Wang et al, 2018), is a collection of tools for evaluating the performance of language models across a diverse set of existing natural language understanding tasks, adopted from different sources. These tasks are divided into two parts: single sentence classification tasks and sentence pair classifications tasks subdivided further into similarity and inference tasks. GLUE also includes a hand-crafted diagnostic test, which probes for complex linguistic phenomena, such as the ability of the model to express lexical semantics and predicate-argument structure, to possess logical apparatus and knowledge representation. GLUE is recognized as a de-facto standart benchmark to evaluate transformer-derived language models. Last but not least, GLUE informs on human baselines for the tasks, so that the submitted models are compared not only to the baseline, but also to the human performance. The SuperGLUE, designed by (Wang et al, 2019), follows GLUE paradigm for language model evaluation, providing with more complex tasks, of which some require reasoning capabilities, and some are aimed at detecting ethical biases.

The choice of training data is one of the most important design choices when training DSMs. Wikipedia, news corpora and web corpora are common sources for DSM training. The corpus size can be estimated by the number of tokens[8]: so, the size of English Wikipedia is 1M tokens, the size of Google news corpus is 1B tokens, and the size of CommonCrawl corpus is 600B tokens. Structured expert-based knowledge bases, such as WordNet, can be used to train word embeddings (Rodrigues et al., 2018) and language models, such as KnowBERT (Peters et al., 2019).

---

[8] According to FastText web page: https://fasttext.cc/docs/en/english-vectors.html

## 4. Brain mapping with DSMs

### 4.1 Count-based DSMs

The pioneering work by (Mitchell et al. 2008) was the first to introduce the task of relating language stimuli with the recorded brain activity. In contrast to primarily descriptive theories, this work presents a first computational model to predict spatial patterns of functional magnetic resonance imaging (fMRI) neural activations associated with previously unseen words. The language stimuli consist of 60 word-picture pairs, the words are concrete nouns and belong to twelve semantic categories. Nine healthy participants took part in the study. To predict activation at voxel $y_v$ a linear model is used: $y_v = \sum_{v_i} c_{v_i} f(w_i)$, where $c_{v_i}$ are trainable weights and $f$ are word vectors. The word vectors are built from word-context co-occurrence statistics. Two types of contexts are defined: either frequent words, according to a large corpus, or the list of manually selected action and sensory-motor verbs. The model is trained in a leave-one-out fashion: all but two words are used for learning model weights, two words are used to test the model precision. The omitted words are chosen either from unrelated categories or from the same category. The verb-based context resulted in a model that significantly outperforms models, based on other contexts. Distinguishing between words within the same category appears to be a more challenging task than predicting words from unrelated categories. The results of the paper show a direct, predictive relationship between the statistics of word co-occurrence and the neural activation associated with thinking about the word meaning and establish a whole new line of research.

Murphy et al. (2012) seems to be one of the next papers that used DSMs based on count-based word embeddings for brain mapping. They used the dataset from Mitchel et al. (2008), containing fMRI data from 9 participants, with stimuli being line-drawings accompanied by their text labels. The authors compared semantic representations that provided state-of-the-art (SOTA) performance on the brain mapping tasks with new distributional models of semantics which could be derived from arbitrary corpora. The SOTA models used were hand-crafted, based on manual annotation, use of domain-appropriate corpus, etc. As competing corpus-derived models the authors considered word-region co-occurrences, word-collocate features including raw tokens, POS tags, etc. To preprocess data and extract features, they used a common pipeline, including a co-occurrence frequency cutoff, a frequency normalisation weighting, and dimensionality reduction with SVD. As a result, the authors claim that they were able to achieve brain activity prediction accuracies comparable to SOTA, or slightly higher, using unsupervised data-driven models of semantics, based on a large random sample of web-text.

## 4.2 Prediction-based DSMs

For over a decade, datasets in neurolinguistics were limited to those, proposed originally e.g. by Mitchel et al. (2008). However as more advanced DSM's emerged, the more applications to neurolinguistics rose. Abnar et. al (2018) modeled brain activity using fMRI data from Mitchel et al. (2008) and different types of word embeddings: experiential, distributional and dependency-based. The main aim was to estimate their usefulness for predicting the neural activation patterns associated with concrete nouns. Here in case of experiential word embeddings 65 features are defined and crowdsourcing is used to rate the relatedness of each feature for each word; as distributed word embeddings they used word2vec, fasttext, dependency-based word2vec, GloVe and LexVec approaches based on ideas similar to ones considered in Section 3 above; dependency-based word embeddings contains 25 verb features and feature-based word vector representations (called "non-distributional" in this paper), where words are presented as binary vectors where each element of the vector indicates whether the repre- sented word has or does not have a specific feature. It turned out that for predicting neural activation patterns in human brains dependency-based word2vec, GloVe and 25 features model are the top-ranked models for at least one of the subjects, and the feature-based word representation model has the lowest performance. At the same time when predicting the word representation given a brain activation, they found that for the 25 features model, the accuracy on the reversed task is much lower. On the other hand, it seems that it is very easy to construct GloVe word vectors from brain activation patterns. Hence, obtaining a high accuracy in the word prediction task using GloVe, supports the fact that the context of the words has a major role in the way we learn the meanings of the words. Besides that, the authors also found out that different models make different kinds of mistakes and identified the most predictable voxels in the brain for each word embedding model.

Rodrigues et al. (2018) went further in their research and instead of using feature-based models or semantic spaces (aka word embeddings) for word representation they utilized a semantic network to encode lexical semantics. Here the semantic network was used as the base repository of lexical semantic knowledge, namely WordNet. The authors then generated semantic space embeddings from semantic networks, and use it to obtain WordNet embeddings: the intuition is that the larger the number of paths and the shorter paths connecting any two nodes (words) in a network the stronger is their semantic association. The proposed model turned out to demonstrate a competitive performance as its accuracy is comparable to the results obtained with SOTA models based on corpus-based work embeddings reported in previous papers. Although for one third of the 9 subjects this model surpasses Mitchell et al. (2008), on average it did not outperform their model based on hand-selected features.

In their seminal paper Huth et al. (2016) consider another type of stimuli, namely, they model brain responses from naturally spoken narrative stories that contain many different semantic domains and map semantic selectivity across the cortex using voxel-wise modelling of functional MRI (fMRI) data. They used a word embedding space to identify semantic features of each word in the stories constructed by computing the normalized co-occurrence between each word and a set of 985 common English words across a large corpus of English text. Constructed fit models are capable of determining which specific semantic domains are represented in each voxel. Due to high dimensionality examining each voxel separately is unfeasible, and so the authors reduced the space of the models by principal component analysis and finally identified four shared dimensions that capture shared aspects of semantic tuning across the subjects and provide a way to summarize succinctly the semantic selectivity of a voxel. To visualize the semantic space, the authors projected the 10,470 words in the stories from the word embedding space onto each dimension and used k-means clustering to identify 12 distinct categories. Thanks to this the authors visualized the pattern of semantic-domain selectivity across the entire cortex by projecting voxel-wise models onto the shared semantic dimensions. In this way for each voxel it is possible to say to which categories in the semantic space greater observed BOLD responses correspond to. Using specially designed probabilistic models and the constructed semantic labelling of voxels in response to stimuli, the authors created a single atlas that describes the distribution of semantically selective functional areas in the human cerebral cortex.

The studies discussed above are limited in the sense that they use relatively small and/or constrained sets of stimuli. Moreover, often the analysis is based on semantic features limited to concrete nouns that do not easily scale to a typical vocabulary of tens of thousands of words. As a result, an important challenge arises whether the models would generalise to meanings beyond the limited scope of the training sample. Pereira et al. (2018) proposed an answer to this challenging question. They developed a system that would word on imaging data collected while a subject reads naturalistic linguistic stimuli on potentially any topic. To represent the semantic space, they used 300-dimensional GloVe vectors and spectral clustering to group words into 200 regions almost all of which were intuitively interpretable. The authors hand-selected representative words from each of the regions and used them in creating the stimuli. Based on collected data Pereira et al. (2018) constructed a universal brain decoder that can infer the meanings of words, phrases, or sentences from patterns of brain activation data. In the first set of experiments they evaluated a capability of single concept decoding. To that the subject is thinking about the relevant meaning of each words, first, the target word was presented in the context of a sentence that made the relevant meaning salient, second, the target word was presented with a picture that depicted some aspect(s) of the relevant meaning, and third, the target word was presented in a "cloud", surrounded by five representative words from the semantic cluster. To evaluate decoding performance Pereira et al. (2018) considered a pairwise classification approach and a rank accuracy classification task. In the former they computed and compared the similarity between the decoded vectors and the "true" (text-derived) semantic vectors. In the latter they compared each decoded vector to all 180 text-derived semantic vectors and ranked them by their similarity. In both cases classification accuracy turned out to be around 0.7 or more. In other experiments on sentence decoding the decoder was applied to the brain images for the sentences from yielding a semantic vector for each sentence. A text-derived semantic vector for each sentence was created by averaging the corresponding word embeddings vectors. To evaluate accuracy Pereira et al. (2018) considered a pairwise classification task to recognise different topics, different passages within the same topic and different sentences from the same passage for all possible pairs in each task. It turned out that it is possible to distinguish sentences at all levels of granularity. As a result, this work clearly demonstrates that it is possible to extract a linguistic meaning from imaging data.

As we see, prediction of semantic meaning of words based on fMRI data prevails. So, it is not surprising that Fyshe et al. (2019) recently concentrated on MEG data and analysed semantic representations of adjective-noun phrases. For that they presented phrases consisting of an adjective followed by a noun. Besides that, some additional unexpected phrases were demonstrated, such as adjective-adjective pairs. The word vector model used was built using a large corpus of text from web pages. The probability of seeing two words in a particular dependency relationship in a sentence was calculated. To get word embeddings the authors compressed the matrix with the probabilities using singular value decomposition. The first 100 dimensions of the matrix were used for this study. Thanks to the experimental setup, there is a strong correlation between adjectives and nouns in the collected data. Two tasks of predicting the stimulus from the MEG recording were considered: 1) predict adjective semantics: predict the identity of the first word in the
phrase (any of the six adjectives); 2) predict noun semantics: predict the
identity of the noun (one of eight). Their analysis revealed two novel
findings: (a) a neural representation of the adjective is present during
noun presentation, but this representation is different from that observed
during adjective presentation and (b) the neural representation of
adjective semantics observed during adjective reading is reactivated after
phrase reading, with remarkable consistency.

## 4.3 Deep contextualized DSMs

As we discussed above the previous approaches mapped word-level
semantic representations using embedding vectors and neglected the
effect of context by assuming that the response to each word is
independent. In this section, applications of pretrained LM's, otherwise
contextualized DSMs, are presented.

Prior to recent advances in language model pre-training, Jain et al.
(2018) used representations discovered by a long-short term memory
network (LSTM) to incorporate context into encoding language models
that predict fMRI responses to natural, narrative speech on the same
dataset as Huth et al. (2016). These contextualized models perform
significantly better at predicting brain responses than previously
published word embedding models. Moreover, Jain et al. (2018)
examined which brain areas can be best modelled by which context
length and layers of the LSTM encoding model. It turned out that voxels
in low-level language areas (AC) prefer short context, while voxels in
higher-level language areas prefer long context. As for layers the authors
showed that there is little difference in performance from different layers,
although layers 1 and 3 are better suited to model AC, layer 2 better
models higher semantic regions. The results illustrate a temporal
hierarchy in how the human language is processed.

Schwartz and Mitchell, (2019) utilize ULMFiT (Howard and Ruder,
2018), a bidirectional multi-layer LSTM, to predict several time-locked
stereotyped EEG responses, known as event-related potentials, ERPs to
word presentations, which are supposed to be markers for semantic or syntactic processes that take place during comprehension. In this study the ERP data comes from the project of Frank et al., (2019). The context-aware word embeddings from each layer of the architecture are concatenated to achieve a single representation to each word in the sentence, fed into a convolution layer and a non-linear layer. The resulting word embeddings are combined with the log-probability of a word and the word-length to a prediction of each ERP. A single loss might be used to train the whole model to predict six ERP signals (or any subset of them) simultaneously. Such multitask learning allows to find relationships between ERP components: the prediction of one ERP component may benefit from inclusion of another ERP component or multiple components. Shwartz and Mitchel, (2019) claim that their results suggest that 8 pairs of ERP signals are related to each other: the LAN is paired with the P600, EPNP, and PNP, the ELAN with the N400, EPNP, PNP, and P600, and the EPNP is paired with the P600. The ablation study shows that the model benefits significantly from the usage of bidirectional architecture. A possible explanation for this may be that the context-aware embeddings benefit from seeing the future language input. Clearly, the model has no access to future brain-activity, so there is no overfitting to the training data. What is more, the bidirectional model has advantageously more parameters, which helps to gain more knowledge from the training data.

Hollestein et al., (2019) present a new framework, Cognival, and conduct an extensive comparison of how different types of DSMs, such as word embeddings and universal language models, fit to various datasets of eye-tracking, EEG and fMRI. They define a single model, namely, neural architecture regression network with one hidden layer, which is used multiple times to predict brain activity from DSMs. The word embeddings under consideration are GloVe, word2vec, fasttext, as well as a less known modification of word2vec, wordnet2vec (Bartusiak et al., 2019), which utilizes WordNet instead of raw textual data. These are compared to two pre-pretrained LMs, namely, BERT and ELMo. The results show that BERT, ELMo and FastText embeddings achieve the best prediction results for predicting all types of brain activity. Universal LMs, otherwise addressed as deep contextualized DSMs, BERT and ELMo, significantly outperform static word embeddings. This supports the general trend in NLP. There is an evident correlation between predictions of eye-tracking, EEG and fMRI, which supports the idea that the DSMs predict actually brain activity signals and not overfit to preprocessing artifacts of each modality.

In contrast to other research projects, Toneva and Wehbe, (2019) do not predict the brain activity from DSMs, but rather use the brain activity recordings as a proxy for interpreting neural DSMs.

The fMRI and MEG data published by Wehbe et al. (2014b) is used as a source for information for interpreting four universal LMs, ELMo, USE, BERT and Transformer-XL. The authors train a linear model with a ridge penalty to predict brain activity from a sentence embedding and show that the middle layers of all DSMs outperform lower and upper layers at the same time. The performance of ELMo, BERT and Transformer-XL is almost the same, with USE having lower performance. Transformer-XL appears to benefit from larger size of contexts, as it is the only DSM specially designed for processing long sequences.

As we can see, there exists various setups of experiments on brain mapping, utilizing different data modalities and types of unsupervised word/sentence embeddings. Obtained results clearly demonstrate the possibility of constructing a brain decoder that is capable of decoding semantics from brain activation data.

## 5. Brain-aware embeddings

Multimodal DSMs combine word and sentence representations with other modalities, such as images and audio recordings, to learn joint word-image embeddings (Mao et al., 2016) or acoustic word embeddings (Chung et al. 2017). Although multimodal DSMs are demanded in NLP, little progress has been achieved in the development of brain-aware embeddings, i.e. such DSMs that combine word stimuli with MRI recordings. At the core of the concept of brain aware embeddings is the following hypothesis: if brain activation data encodes semantics, inclusion of brain data in a model of semantics results in a model more consistent with semantic ground truth (Fyshe et al., 2014). A first attempt to build a brain-aware DSM was made by (Rustandi et al., 2009).

Canonical Correlation Analysis (CCA) was applied to correlate corpus-based and brain-based statistics. However, as CCA requires that the data sources be paired, a vast majority of the corpus-bases statistics was lost, thus the benefits of CCA were very limited, as there was little overlap between stimuli words and words in the corpus with the former being the bottleneck.

Another of the first models of brain-aware embeddings is introduced by Fyshe et al., (2014). The proposed model, Joint Non-Negative Sparse Embedding (JNNSE), utilizes matrix factorization techniques to build a hybrid count-based DSM. JNNSE solves the following objective: $argmin_{A,D^c,D^b}||X - AD^c|| + ||Y - AD^b|| + \lambda||A||_1$, where $X$, $Y$ are corpus-based and brain-based word matrices. $X$ represents standard word-context co-occurrence, while $Y$ is constructed from feature vectors derived from brain activity recording. The JNNSE task is solved with respect matrices $A, D^c, D^b$ being non-negative. The results show that even a small available amount of brain data has a positive impact on the embedding, as they are better in predicting fMRI activations, than text-based embeddings.

Schwartz et al., (2019) have recently introduced the first model specifically designed to capture the way the brain represents language meaning. They fine-tuned the BERT model (Devlen et al., 2018) to predict recordings of brain activity of people, while reading a text. The resulting representations are encoding more brain-activity-relevant information and thus improve the quality of the brain activity prediction. To make sure that the fine-tuned BERT is not overfitted for the task, the authors of the paper conduct a series of transfer learning experiments and show the following. First, the changes to language representation does not harm the performance on downstream NLP tasks, when tested on GLUE benchmark. Second, the model is capable of generalizing to new participants, i.e. the model does not overfit on training data. Third, the changes introduced to the model are not an artifact of imaging modality, as the fine-tuned representations learned from both MEG and fMRI are better for predicting fMRI than predicting from fMRI-based representations alone.

To encode the brain activity information, the common procedure for fine-tuning BERT is used: a simple linear layer is added on top of the universal language model. This linear layer is trained to map the output embeddings from the base architecture to the fMRI prediction task. The added linear layer can be either trained alone or the whole model can be fine-tuned in the end-to-end fashion. Two datasets of MEG and fMRI experiments (Wehbe et al. 2014a, Wehbe et al. 2014b) were used to fine-tune the pre-trained BERT model by (Devlin et al., 2018).

As fMRI and MEG recordings have different time resolution, they are used differently to supervise the fine-tuning procedure. To predict fMRI, the pooled embedding of the sentence is used, while the MEG recordings are used in a sequence labelling fashion, so that each word in the input sequence is mapped to a certain MEG recording.

To demonstrate whether the model can be biased towards brain-relevant language information, several models are trained. A vanilla model, which serves as a baseline, is trained for each experiment participant in such a way, that only an added linear layer is updated, and the core layers of the model remain unchanged. In contrast to the vanilla model, the fully joint model is trained to simultaneously predict all participants' data from both fMRI and MEG experiments. The results of the paper show that the joint model performs better than the vanilla model. This means that the model poses the ability to generalize across participants and imaging modalities. Although the performance of the joint model on such NLP tasks as GLUE benchmark does not differ much from the initial BERT model. It remains unclear how exactly the language representations were changed during the fine-tuning procedure. Finally, this study demonstrated a first step towards brain activity aware language models.

## 6. Medical applications

Many approaches taken from neurolinguistics, DSM in particular, found applications in medical studies of neurological and psychiatric disorders, like Mild Cognitive Impairment (MCI), aphasia and others. Even without sophisticated neuroimaging techniques, word embeddings models could help in clinical diagnostics and prognosis tasks. Many studies confirm that Semantic Verbal Fluency (SVF, a task where patients should recall as many words from a particular semantic category as they can in a short period of time) is helpful for diagnostics of such brain pathologies as Alzheimer's disease (Pakhomov et al., 2016), MCI, focal brain lesions (Troyer et al., 1998), as well as schizophrenia (Robert et al., 1998). Many studies explored a machine learning approach to automatically detect cognitive impairment based on text/speech samples from ill subjects and healthy controls. In a paper (Linz et al., 2017) the authors compared the classification performance of traditional taxonomic models (Troyer et al., 1997) with word2vec model for MCI detection on a sample of 100 subjects and their results were clearly in favour of the DSM. Another work (Fraser et al., 2019) also suggests that the fully automated cluster models are better able to capture patterns in the data which distinguish the MCI patients from healthy controls. In this study authors extract information units using topic models trained on word embeddings in monolingual and multilingual spaces and find that the multilingual approach leads to significantly better classification accuracies than training on the target language alone. This might be due to the fact that multilingual data enriches the vocabulary in the relevant areas of the semantic space and topic modelling approach helps to better identify this space (Fraser et al., 2019).   Another example is clinical assessment of aphasia using a confrontation naming task, where a stimulus is shown to a subject and she must name it (McKinney-Bock and Bedrick, 2019). This procedure is very time demanding and need automation in assessing semantic similarity. The authors in (Mckinney-bock and Bedrick, 2019) as well as (Fergadiotis et al., 2016) showed that word embeddings can be successfully applied to classification of aphasias when assessed by the naming task.

When a person performs either speech generation or comprehension, it is always connected networks of different brain regions working together to perform such a task. For example, the fronto-temporal language system has been shown to respond very specifically in fMRI to linguistic stimuli, but not to other perception or cognitive tasks (Ivanova et al., 2019). One question is whether the language system is engaged during the processing of words meaning and is necessary for it. In the paper (Ivanova et al., 2019) fMRI study on patients with brain damage was combined with neurolinguistics approach - distributed semantic representations created through extensive training on massive language corpora. The authors found that language brain networks in healthy controls are active during the semantic plausibility task when subjects were shown both sentences and pictures. In patients with global aphasia (those having language areas severely damaged), performance on the picture plausibility task was not decreased, suggesting that the language network is not required for understanding and accessing semantic information in the brain about visually presented events. Much more information about possible mechanisms of brain disorders could be captured when analyzing the same semantic models with different brain imaging modalities. A fresh example is in the paper (Hollenstein et al., 2019), where the authors present the first multimodal framework for evaluating English word representations based on cognitive lexical semantics. They investigated six types of word embeddings by fitting them to 15 datasets of eye-tracking, EEG and fMRI data collected during language processing tasks. A remarkable result here is that the authors found correlations in the embedding performance between different cognitive datasets (i.e. different tasks and different neuroimaging modalities) and their performance on extrinsic NLP tasks. Moreover, the authors developed a framework to automate these comparisons. Though the authors demonstrate framework effectiveness on healthy cohort, it is applicable to medical studies. Thus, neurolinguistic experiments with neuroimaging data collection could give insights to applied NLP tasks and new word embedding creation and evaluation.

Linguistic models might also give insights on aphasia treatment. The authors in (Roelofs, 2019) explore the method called phonological cueing to increase the word retrieval (or proper word finding) performance in patients with aphasia. Picture naming performance of the WEAVER computer model (Roelofs, 1997) was assessed in simulations of immediate and treatment effects of phonological cueing in post-stroke aphasia. The authors claim that the linguistic model successfully simulated the observed effects in naming performance and neural measures. The model could expand our understanding of word finding, associated difficulties, and their improvement by therapy. As the WEAVER model is initially a theoretical model of the word-form encoding process, it benefits from adding empirical data, including physiological and patient data (Roelofs, 2019).

The field of neurolinguistics and semantic models can also help the area, closely related to brain mapping Direct-Speech Brain-Computer-Interface (DS-BCI) research on methods for targeting the imagined speech content for synthesis (Cooney et al., 2018). DS-BCI is aimed at acquiring neural signals (mostly EEG and EcoG due to high temporal resolution) corresponding to imagined speech in order to decode these signals and produce a linguistic output in the form of phonemes, words, or sentences (Cooney et al., 2018) to help disabled people and thus could be considered as an inverse problem of brain mapping. Recent studies have shown the potential of neurolinguistics to improve the ability of DS-BCI to decode imaginary speech from acquired signals with the inclusion of semantics in experimental procedures.

## 7. Conclusion

The rapid growth of interest towards neurolinguistics research across scientific teams led to an increase in the number of in-house, multimodal datasets with different imaging modalities, different linguistic tasks and types of stimuli. This wide variety of datasets makes it difficult to compare results from different teams or re-use these data in new brain mapping studies. The possible way out for prospective studies could be to standardize (to some extent) the imaging procedure and at least most common linguistic tasks, as it was successfully done previously in the field of machine learning, for instance, in well-known image classification databases, or more general brain mapping experiments, see Human Connectome Project (Van Essen et al. 2013) and similar. In order to make comparisons across studies and perform meta-analysis, we should elaborate on some common performance (or accuracy) measures for brain mapping experiments.

As already mentioned, up to now there is no unified approach or standardization on how to carry out neurolinguistics experiments and acquire brain imaging data, so lots of heterogeneous multimodal datasets already exist, which could not be processed and analyzed in the same way. Moreover, usually these datasets are quite small in terms of number of subjects, see for example (Hanke et al. 2014; Huth et al. 2016; Wehbe et al. 2014) where there are less than 10 subjects in the dataset. In order to be able to use the data from already existing different datasets for models training or testing, it is necessary to develop new transfer learning approaches which take into account intersubject variability as well as scanning on different brain imaging equipment (with different settings) and using different experimental paradigms.

From our perspective, we would like to point out the following three directions of future research and development. First of all, we foresee the advances of multimodal models, which blend different types of language and neuroimaging data, measured in different time scales. This aligns both with the current trends in both natural language processing and neuroimaging. As noted, most of the studies discussed are English-centered. Therefore, the second future development direction covers multilingual and cross-lingual aspects. It is essential to study whether the universal language models show the same patterns for different languages and develop methods to compare experimental results for different languages. Finally, we need to acknowledge the shortcomings of data-driven approaches such as the absence of solid theoretical grounding. Experiments suggest only a correlation between language model prediction and measured brain activity, though there is little theoretical understanding of how a human brain processes semantics. This limits the medical applications significantly. The third direction of future work concentrates on the development of the necessary theoretical framework.

## 8. Acknowledgements

The authors would like to thank the anonymous reviewers for valuable comments and suggestions. Ekaterina Artemova was supported by the framework of the HSE University Basic Research Program and Russian Academic Excellence Project "5-100", Maxim Sharaev was supported by Russian Foundation for Basic Research according to the research project №17-29-02518 (medical applications of neurolinguistics).

 Author